# Compatibility of Quantitative and Qualitative Representations of Belief

S.K.M. Wong, Y.Y. Yao, and P. Lingras
Department of Computer Science, University of Regina
Regina, Saskatchewan, Canada S4S 0A2

## Abstract

The compatibility of quantitative and qualitative representations of beliefs was studied extensively in probability theory. It is only recently that this important topic is considered in the context of belief functions. In this paper, the compatibility of various quantitative belief measures and qualitative belief structures is investigated. Four classes of belief measures considered are: the probability function, the monotonic belief function, Shafer's belief function, and Smets' generalized belief function. The analysis of their individual compatibility with different belief structures not only provides a sound basis for these quantitative measures, but also alleviates some of the difficulties in the acquisition and interpretation of numeric belief numbers. It is shown that the structure of *qualitative probability* is compatible with monotonic belief functions. Moreover, a belief structure slightly weaker than that of *qualitative belief* is compatible with Smets' generalized belief functions.

## 1 INTRODUCTION

Uncertainty is always present in modeling realistic situations. It may stem from a lack of knowledge, the incompleteness or the unreliability of the information at our disposal. In order to draw a meaningful conclusion under uncertain situations, we may have to express our beliefs in a number of propositions. Many approaches have been proposed for representing, measuring, and reasoning with uncertain information. Despite the diversities of these methods, one can divide them into two classes: the quantitative (numeric) and the qualitative (non-numeric) approaches (Bhatnagar and and Kanal, 1986; Spiegelhalter, 1986; Satoh, 1989). In the quantitative approach, a number is associated with each proposition to indicate the degree to which one believes in that proposition. That is, we express our belief in a proposition by a numeric value. To make the quantitative representation of beliefs consistent and meaningful, certain axioms or rules should be observed in expressing one's beliefs. For example, if beliefs are measured by a probability function, the Kolmogorov axioms for probability should be satisfied in order to maintain consistency. In the qualitative approach, beliefs are expressed by a preference relation on a set of propositions. As in quantitative measures of beliefs, such a relation must be consistently defined. For example, if a person believes more in proposition $A$ than in proposition $B$, and also believes more in $B$ than in $C$, then it is reasonable to assume that he would believe more in $A$ than in $C$. In this paper, we are interested in those belief structures which are compatible with some well known belief measures.

Both the quantitative and qualitative approaches are very useful for the management of uncertainty. In fact, probability theory has been extensively studied within the quantitative as well as the qualitative frameworks (Fishburn, 1970; Savage, 1972; Fine, 1973). Given a belief measure and a preference relation, an important question one inevitably would ask is whether they are compatible with each other. This is indeed one of the fundamental issues in measurement theory, which is concerned to a large extent with the mathematical modeling of preferences and beliefs (French, 1986). Depending on the context, a preference relation is also referred to as a comparative probability, possibility, or belief relation. The compatibility of a comparative probability relation and a probability function was investigated by many authors (Fishburn, 1970; Savage, 1972; Fine, 1973). Dubois (1986) studied the compatibility of a comparative possibility relation and a possibility function. Possibility functions were originally proposed by Zadeh (1978) within the framework of fuzzy sets, and they were later shown to be closely related to consonant belief functions introduced in the theory of evidence (Shafer, 1976). Wong et al. (1990) studied the compatibility of a comparative belief relation and a belief function. There is an important class of preference relations referred to as qualitative probability (Savage, 1972). However, until now it is not known which class of belief functions is compatible with qualitative probability relations. Recently, Smets (1988) proposed a generalized version of belief functions. It is interesting to investigate what kind of preference relation is compatible with Smets' generalized belief functions.

In this paper, our discussion will focus on the compatibility of quantitative and qualitative representations of beliefs. In particular, we analyze four classes of quantitative belief measures, namely, the probability function, the monotonic belief function, Shafer's belief function, and Smets' generalized belief function, and their compatibility with different kinds of preference relations. We will show that qualitative probability relations are compatible with monotonic belief functions, and that a preference structure slightly weaker than that of qualitative belief is compatible with generalized belief functions. More importantly, we believe that the study of the compatibility of these qualitative and quantitative representations of beliefs may provide a foundation for developing a generalized utility theory (Jaffray, 1989).

## 2 QUANTITATIVE BELIEF MEASURES

Based on the notion of belief functions (Shafer, 1976), we will identify four different classes of quantitative measures of belief.

Let $\Theta = \{\theta_1, \ldots, \theta_s\}$ denote a finite set of possible answers to a question, which is referred to as the *frame of discernment* or simply the *frame* defined by the question. Following the convention of representing a proposition by a subset of $\Theta$, the power set $2^\Theta$ denotes the set of all propositions discerned by frame $\Theta$. A quantitative belief measure can be viewed as a mapping from $2^\Theta$ to the real numbers.

**Definition 1:** A probability function $P$ is a mapping from $2^\Theta$ to the interval $[0,1]$, $P : 2^\Theta \to [0,1]$, which satisfies the following axioms:

(B1) $P(\emptyset) = 0$,

(B2) $P(\Theta) = 1$,

(B3) For $A, B \in 2^\Theta$ with $A \cap B = \emptyset$,
$P(A \cup B) = P(A) + P(B)$.

Axiom (B3) is usually referred to as the *additivity* axiom. By replacing this axiom with the *sup-additive* axiom, another class of quantitative belief measures called belief functions (Shafer, 1976) can be defined as follows.

**Definition 2:** A belief function $Bel$ is a mapping from $2^\Theta$ to the interval $[0,1]$, $Bel : 2^\Theta \to [0,1]$, which satisfies (B1), (B2), and the sup-additive axiom:

(B3') For every integer $n > 0$ and every collection $A_1, A_2, \ldots, A_n \in 2^\Theta$,
$$Bel(A_1 \cup A_2 \ldots \cup A_n) \geq$$
$$\sum_i Bel(A_i) - \sum_{i<j} Bel(A_i \cap A_j) \pm \ldots +$$
$$(-1)^{n+1} Bel(A_1 \cap A_2 \ldots \cap A_n).$$

A belief function can be equivalently defined by a mapping from $2^\Theta$ to the interval $[0,1]$, $m : 2^\Theta \to [0,1]$, which is called a basic probability assignment satisfying the axioms:

(M1) $m(\emptyset) = 0$,

(M2) $\sum_{A \in 2^\Theta} m(A) = 1$.

In terms of the basic probability assignment, the belief in a proposition $A \in 2^\Theta$ can be expressed as:

(M3) $Bel(A) = \sum_{B \subseteq A} m(B)$,

where the summation is restricted to the elements $B$ in $2^\Theta$, which are subsets of $A$. Conversely, given a belief function one can construct the corresponding basic probability assignment. Therefore, belief functions can be defined either by axioms (B1), (B2), and (B3') or by axioms (M1)-(M3).

Axioms (B1) and (B2) state that the proposition $\emptyset$ is believed to be false (impossibility) and the proposition $\Theta$ is believed to be true (certainty). These two axioms indicate that the *closed* world assumption (Smets, 1988) is in fact used to define the frame of discernment $\Theta$. That is, one has implicitly assumed that the frame $\Theta$ consists of *all* possible answers to a given question and only one of these answers is correct. On the other hand, axiom (B3') indicates that belief functions are sup-additive, and become additive in the degenerated case. Thus, additive probability functions belong to the class of degenerated belief functions.

Note that the additivity axiom implies the monotonicity axiom, namely:

(B4) For $A, B, C \in 2^\Theta$ with $(A \cup B) \cap C = \emptyset$,
$P(A) > P(B) \iff P(A \cup C) > P(B \cup C)$.

However, monotonicity does not imply additivity, and axioms (B1), (B2), and (B3') do not imply monotonicity. This means that belief functions do not necessarily satisfy axiom (B4). In some applications, it is desirable that the monotonicity axiom (B4) is satisfied (Savage, 1972). In that case, we can define another class of belief measures, which falls between the belief functions and the probability functions.

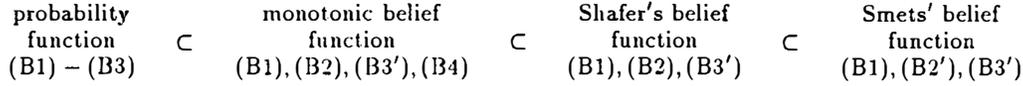

Figure 1: Relationships Between Quantitative Measures of Belief

**Definition 3:** A belief function $Bel$ is called a *monotonic* belief function if it satisfies the monotonicity axiom (B4).

We will show that monotonic belief functions are compatible with the qualitative probability relations.

Smets (1988) pointed out that the condition (M1), $m(\emptyset) = 0$, reflects the *closed* world assumption implicitly used in Shafer's definition of belief function; $m(\emptyset) > 0$ reflects the *open* world assumption. With the open world assumption, the probability mass $m(\emptyset)$ can be interpreted as the belief committed exactly to the proposition that the true answer is not in the frame $\Theta$. Smets' belief functions, written $bel$, satisfy the following axioms:

$$(M2) \quad \sum_{A \in 2^\Theta} m(A) = 1,$$

$$(M3') \quad bel(A) = \sum_{\substack{B \subseteq A \\ B \neq \emptyset}} m(B).$$

The condition $m(\emptyset) > 0$ has also been considered by Dubois and Prade (1986) under a set-theoretic view of belief functions. The generalized belief functions can be equivalently defined as follows.

**Definition 4:** A generalized belief function $bel$ is a mapping from $2^\Theta$ to the interval $[0, 1]$, $bel : 2^\Theta \to [0, 1]$, which satisfies axioms (B1), (B3'), and

$$(B2') \quad bel(\Theta) = 1 - m(\emptyset),$$

where $0 \leq m(\emptyset) \leq 1$.

In the above discussion, we have considered four classes of quantitative belief measures: the probability function, the monotonic belief function, Shafer's belief function, and Smets' generalized belief function. The relationships between these measures are shown in Figure 1. The set of probability functions is a subset of the set of monotonic belief functions, and so on. In the following section, we will study the preference structures that are compatible with these quantitative belief measures.

## 3 PREFERENCE RELATIONS VERSUS QUANTITATIVE BELIEF MEASURES

In the qualitative representation of beliefs, it is assumed that one is able to express one's preference on any two propositions $A, B \in 2^\Theta$, without stating numerically how much one prefers proposition $A$ to proposition $B$. Qualitative judgments can be described in terms of a preference relation $\succ$. By $A \succ B$, we mean that $A$ *is preferred to* $B$. In the absence of strict preference, i.e., $\neg(A \succ B)$ and $\neg(B \succ A)$, we say that $A$ and $B$ are *indifferent*, written $A \sim B$. We also write $A \succeq B$ if $A \succ B$ or $A \sim B$. The relationship between the quantitative and qualitative representations of beliefs can be formally stated as follows.

**Definition 5:** Suppose $\Theta$ is a frame, $f$ is a function mapping the elements of $2^\Theta$ onto the set of real numbers, and $\succ$ is a preference relation on $2^\Theta$. We say that $f$ and $\succ$ are *compatible* with each other if for $A, B \in 2^\Theta$,

$$A \succ B \iff f(A) > f(B).$$

A function $f$ is said to *represent* $\succ$ if it is compatible with $\succ$.

Clearly, whether a preference relation is compatible with a particular quantitative belief measure depends very much on the preference structure representing the qualitative judgments.

Now consider a special class of preference relations characterized by the following two axioms:

(Q1) asymmetric :
$$A \succ B \implies \neg(B \succ A),$$

(Q2) negatively transitive :
$$(\neg(A \succ B), \neg(B \succ C)) \implies \neg(A \succ C).$$

Axiom (Q1) suggests that if one commits more belief in $A$ than in $B$, one should not at the same time commit more belief in $B$ than in $A$. If this axiom holds for a preference relation $\succ$, then for every $A, B \in 2^\Theta$, $A \succeq B \iff \neg(B \succ A)$. Axiom (Q2) demands that if one does not commit more belief in $A$ than in $B$, nor commits more belief in $B$ than in $C$, one should not commit more belief in $A$ than in $C$. A preference relation $\succ$ satisfying these two axioms is called a weak order which can be represented by a real-valued function (Fishburn, 1970; Roberts, 1976).

**Theorem 1.** Suppose $\Theta$ is a finite set and $\succ$ a preference relation on $2^\Theta$. There exists a real-valued function $f$ on $2^\Theta$ such that for every $A, B \in 2^\Theta$,

$$A \succ B \iff f(A) > f(B)$$

if and only if the relation $\succ$ satisfies axioms (Q1) and (Q2). Moreover, $f$ is uniquely defined up to a strictly monotonic transformation.

This theorem is important because it suggests that any belief characterized by a weak order can be measured in terms of an ordinal scale. Theorem 1 therefore provides a basis for representing various types of preference relations. However, axioms (Q1) and (Q2) alone are not sufficient to guarantee that the preference structure is compatible with any of the belief measures introduced in the last section. Additional conditions are required to differentiate different preference structures.

A special type of preference relation known as *qualitative probability* was studied extensively in probability theory (de Finetti, 1937; Fishburn, 1970; Savage, 1972; Dubois, 1986).

**Definition 6:** Let $\Theta$ be a frame. A preference relation $\succ$ defined on $2^\Theta$ is called a *qualitative probability* relation if it satisfies (Q1), (Q2) and the following additional axioms: for $A, B, C \in 2^\Theta$,

(Q3)  nontriviality : $\Theta \succ \emptyset$,
(Q4)  improbability of impossibility : $\neg(\emptyset \succ A)$,
(Q5)  monotonicity :
$$(A \cup B) \cap C = \emptyset \implies$$
$$(A \succ B \iff A \cup C \succ B \cup C).$$

Axioms (Q1)-(Q5) are necessary but not sufficient to guarantee the existence of a probability function (Kraft, Pratt, and Seidenberg, 1959). Scott (1964) gave the necessary and sufficient conditions for the existence of a probability function for a finite set. Let $\mu_A$ denote the *characteristic function* of a subset $A$ of $\Theta$ such that $\mu_A(\theta) = 1$ if $\theta \in A$, and $\mu_A(\theta) = 0$ otherwise. Scott's theorem can be stated as follows.

**Theorem 2.** Let $\Theta$ be a frame and $\succ$ a preference relation on $2^\Theta$. There exists a probability function, $P : 2^\Theta \to [0, 1]$, satisfying: for $A, B \in 2^\Theta$,

$$A \succ B \iff P(A) > P(B),$$

if and only if $\succ$ satisfies (Q1), (Q3), (Q4) and the following axiom:

(S)  For all subsets $A_0, \ldots, A_n, B_0, \ldots, B_n$ of $\Theta$,

if $A_i \succeq B_i$ for $0 \leq i < n$, and
$$\mu_{A_0}(\theta) + \ldots + \mu_{A_n}(\theta) = \mu_{B_0}(\theta) + \ldots + \mu_{B_n}(\theta),$$
for all $\theta \in \Theta$, then $B_n \succeq A_n$.

Axiom (S) requires that any element $\theta$ of $\Theta$ is in exactly as many $A_i$ as $B_i$. In fact, axiom (S) implies both axioms (Q2) and (Q5) provided that axiom (Q1) holds. For example, let $(A \cup B) \cap C = \emptyset$. Suppose $B \succeq A$. We have: for all $\theta \in \Theta$,

$$\mu_B(\theta) + \mu_{A \cup C}(\theta) = \mu_A(\theta) + \mu_{B \cup C}(\theta).$$

According to axiom (S), $B \cup C \succeq A \cup C$. Similarly, if $B \cup C \succeq A \cup C$, axiom (S) implies $B \succeq A$. Therefore,

$$(A \cup B) \cap C = \emptyset \implies (B \succeq A \iff B \cup C \succeq A \cup C).$$

Recall that axiom (Q1) implies $B \succeq A \iff \neg(A \succ B)$. Thus,

$$(A \cup B) \cap C = \emptyset \implies (A \succ B \iff A \cup C \succ B \cup C).$$

This means that axiom (S) implies axiom (Q5) if axiom (Q1) holds.

Theorem 2 only suggests the existence of a probability function; there may exist functions other than the probability functions, which are also compatible with a preference relation satisfying (Q1), (Q3), (Q4), and (S).

Since probability functions are a special type of belief functions, it is expected that there exists a weaker preference structure for belief functions.

**Definition 7:** Let $\Theta$ be a frame. A preference relation $\succ$ defined on $2^\Theta$ is called a *qualitative belief* relation if it satisfies (Q1)-(Q3), and the axioms: for $A, B, C \in 2^\Theta$,

(Q4')  dominance : $A \supseteq B \implies \neg(B \succ A)$,
(Q5')  partial monotonicity :
$$(A \supset B, A \cap C = \emptyset) \implies$$
$$(A \succ B \implies A \cup C \succ B \cup C).$$

Axiom (Q3) eliminates the trivial preference relation, i.e., $A \sim B$ for all $A, B \in 2^\Theta$. The dominance axiom (Q4') says that one should not commit more belief in a subset than in the set itself. This axiom is stronger than (Q4). Given the axioms (Q1) and (Q2) of a weak order, the dominance axiom can be expressed equivalently as $A \supseteq B \implies A \succeq B$. Obviously, axiom (Q5') is a weaker form of the monotonicity axiom (Q5). It is important to note that (Q1)-(Q3) together with (Q4')-(Q5') form a set of independent axioms which completely characterize the qualitative belief relations. The following theorem (Wong et al., 1990) shows that qualitative belief relations are indeed compatible with belief functions.

**Theorem 3.** Let $\Theta$ be a frame and $\succ$ a preference relation on $2^\Theta$. There exists a belief function, $Bel : 2^\Theta \to [0, 1]$, satisfying: for $A, B \in 2^\Theta$,

$$A \succ B \iff Bel(A) > Bel(B)$$

if and only if the preference relation $\succ$ is a qualitative belief relation.

We can prove the *only if* part of the theorem trivially from the properties of belief functions. The *if* part of the theorem can be proved by constructing a belief function compatible with a qualitative belief relation. Axioms (Q1) and (Q2) imply that the induced indifference relation $\sim$ is an equivalence relation (Fishburn, 1970). Since axiom (Q5) holds, based on the relation $\sim$, we can partition $2^\Theta$ into at least two equivalence classes $E_0, \ldots, E_k$ ($k \geq 1$). An equivalence $E_i$ is also denoted as $[A]$ if $A \in E_i$. For example, $E_0$ may be written as $[\emptyset]$ and $E_k$ as $[\Theta]$. First we recursively construct a function $f$ on the equivalence classes as follows:

(i) $f(E_0) = 0$,
(ii) $f(E_{n+1}) = \max\{f'(E_{n+1}), f(E_n) + 1\}$

where

$$f'(E_{n+1}) = \max_{A \in E_{n+1}} \{-\sum_{A \supset B} (-1)^{|A-B|} f([B])\}$$

if for every $A \in E_{n+1}$, $A \supset B \implies B \notin E_{n+1}$; otherwise $f'(E_{n+1}) = f(E_n) + 1$. The symbol $|\cdot|$ denotes the cardinality of a set. The function thus constructed may be considered as an *unnormalized* belief function which satisfies $A \succ B \iff f([A]) > f([B])$ for $A, B \in 2^\Theta$. Based on the function $f$, we can then construct a normalized belief function:

$$Bel(A) = \frac{f([A])}{f([\Theta])}.$$

Theorem 3 shows that if a preference relation is a qualitative belief, i.e., it satisfies axioms (Q1)-(Q3), (Q4'), and (Q5'), then there exists a belief function compatible with the relation. However, these axioms do not guarantee that the class of belief functions is the only kind of functions representing qualitative belief (Smets, 1990). As we mentioned earlier, the same can be said about Scott's theorem. For example, consider a preference relation defined by $\{\theta_1, \theta_2\} \succ \{\theta_2\} \succ \{\theta_1\} \succ \emptyset$. Obviously, this relation satisfies the axioms for qualitative belief as well as those required by Scott's theorem. It can be represented either by a probability function:

$$\begin{aligned} P(\emptyset) &= 0.0, \\ P(\{\theta_1\}) &= 0.4, \\ P(\{\theta_2\}) &= 0.6, \\ P(\{\theta_1, \theta_2\}) &= 1.0, \end{aligned}$$

or by a belief function:

$$\begin{aligned} Bel(\emptyset) &= 0.0, \\ Bel(\{\theta_1\}) &= 0.2, \\ Bel(\{\theta_2\}) &= 0.5, \\ Bel(\{\theta_1, \theta_2\}) &= 1.0. \end{aligned}$$

This relation can also be represented by another function $f$ which is neither a probability function nor a belief function:

$$\begin{aligned} f(\emptyset) &= 0.0, \\ f(\{\theta_1\}) &= 0.6, \\ f(\{\theta_2\}) &= 0.7, \\ f(\{\theta_1, \theta_2\}) &= 1.0. \end{aligned}$$

Certainly, it will be useful if one can define a set of axioms to characterize a class of preference relations that can be represented only by belief functions. This may, however, be a difficult task in general. For a special type of belief functions known as consonant belief functions, Dubois (1986) gave such a set of axioms.

Based on Definition 7 and Theorem 3, we can now show that the qualitative probability relations are in fact compatible with the monotonic belief functions as defined by Definition 3.

**Lemma 1.** Suppose $\Theta$ is a finite set and $\succ$ a preference relation on $2^\Theta$. If $\succ$ is a qualitative probability relation, it is also a qualitative belief relation.

**Proof:** Axioms (Q1), (Q2), and (Q3) are satisfied by both qualitative probability and belief relations. Also, (Q5') is a weaker version of the monotonicity axiom (Q5). Thus, it will suffice to prove that the dominance axiom (Q4') follows from those axioms that define qualitative probability. Suppose $A \supseteq B$. Let $A = B \cup C$ and $B \cap C = \emptyset$. From $\neg(\emptyset \succ C)$ and axiom (Q5), it follows that the dominance axiom holds. Therefore, if a preference relation is a qualitative probability relation, it is also a qualitative belief relation. $\square$

**Theorem 4.** Let $\Theta$ be a frame and $\succ$ a preference relation on $2^\Theta$. There exists a *monotonic* belief function, $Bel : 2^\Theta \to [0, 1]$, satisfying: for $A, B \in 2^\Theta$,

$$A \succ B \iff Bel(A) > Bel(B),$$

if and only if the preference relation $\succ$ is a qualitative probability relation.

**Proof:**

(*only if*) Suppose there exists a monotonic belief function $Bel : 2^\Theta \to [0, 1]$ such that $A \succ B \iff Bel(A) > Bel(B)$. Then, the asymmetric and negatively transitive properties of $\succ$ immediately follow from the properties of the relation $>$ on real numbers. In other words, axioms (Q1) and (Q2) hold. The axiom (Q3) of nontriviality, $\Theta \succ \emptyset$, is implied by axioms (B1) and (B2), because $Bel(\Theta) = 1 > 0 = Bel(\emptyset)$. Axiom (Q4) can be trivially proved from the fact that $Bel(\emptyset) = 0$ and $Bel(A) \geq 0$ for all $A \in 2^\Theta$. Now suppose $A \cap C = B \cap C = \emptyset$. From the assumption that $A \succ B \iff Bel(A) > Bel(B)$ and axiom (B4), we obtain:

$$A \succ B \iff Bel(A) > Bel(B)$$
$$\iff Bel(A \cup C) > Bel(B \cup C)$$
$$\iff A \cup C \succ B \cup C.$$

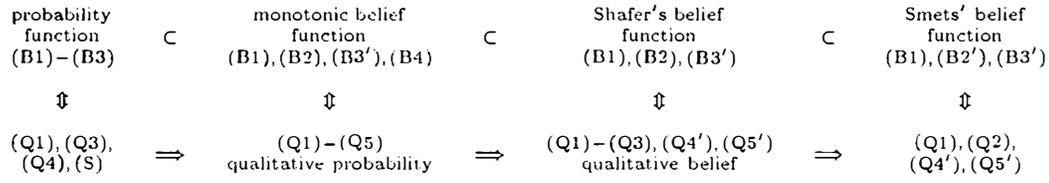

Figure 2: Relationships between Quantitative and Qualitative Representations of Belief

That is, the monotonicity axiom (Q5) holds.

(*if*) From Lemma 1, we know that a qualitative probability relation is also a qualitative belief relation. By Theorem 3, there exists a belief function $Bel$ satisfying the condition: $A \succ B \iff Bel(A) > Bel(B)$ for $A, B \in 2^\Theta$. From the monotonicity axiom (Q5), we have: for $A \cap C = B \cap C = \emptyset$,

$$Bel(A) > Bel(B) \iff A \succ B$$
$$\iff A \cup C \succ B \cup C$$
$$\iff Bel(A \cup C) > Bel(B \cup C).$$

This means that $Bel$ satisfies axiom (B4). □

Note that axioms (B1) and (B2) imply axiom (Q3), i.e., $\Theta \succ \emptyset$. Thus, (Q3) may be weakened or eliminated under the open world assumption. In fact, the following theorem shows that the preference structure compatible with the generalized belief functions can be defined by a set of axioms without (Q3).

**Theorem 5.** Let $\Theta$ be a frame and $\succ$ a preference relation on $2^\Theta$. There exists a generalized belief function, $bel : 2^\Theta \to [0, 1]$, satisfying: for $A, B \in 2^\Theta$,

$$A \succ B \iff bel(A) > bel(B),$$

if and only if the preference relation $\succ$ satisfies axioms (Q1)-(Q2) and (Q4')-(Q5').

**proof:**

(*only if*) The proof is similar to that of the *only if* part in Theorem 4.

(*if*) Since the dominance axiom implies $\neg(\emptyset \succ \Theta)$, we only have to consider two separate cases: (i) $\Theta \sim \emptyset$ and (ii) $\Theta \succ \emptyset$. Obviously, the second case with $\Theta \succ \emptyset$ is equivalent to Theorem 3. If $\Theta \sim \emptyset$, from axioms (Q1), (Q2), and (Q4'), one can immediately conclude that for any $A, B \in 2^\Theta$, the relationship $A \succ B$ is always false, namely, $A \sim B$ is always true (Wong, Bollmann, and Yao, 1990). In this case, we can construct a generalized belief function by letting $m(\emptyset) = 1$. That is, according to (M3'), $bel(A) = 0$ for all $A \in 2^\Theta$. For this belief function $bel$, the condition $A \succ B \iff bel(A) > bel(B)$ holds for any $A, B \in 2^\Theta$. □

The results of the compatibility of preference relations and belief measures are summarized in Figure 2. It can be seen that the inclusion relation $\subset$ between different classes of quantitative belief measures corresponds to the implication relation $\Longrightarrow$ between different sets of axioms defining the various preference relations. The links established here between these belief measures and preference relations provide a better understanding of modeling uncertainty with beliefs.

In this preliminary investigation, we have not considered all the important classes of belief functions. It is worth mentioning here that Dubois (1986) proposed a set of axioms to characterize consonant belief functions. A more detailed analysis of various types of preference relations will be reported in a subsequent paper.

## 4 CONCLUSION

In this paper, we studied the compatibility of quantitative and qualitative representations of beliefs. In particular, four classes of quantitative belief measures were analyzed, namely, the probability function, the monotonic belief function, Shafer's belief function, and Smets' generalized belief function. We established their individual compatibility with different belief structures. These compatibility relationships not only provide a justification for the use of these quantitative measures, but also alleviate some of the difficulties in the acquisition and interpretation of numeric belief numbers.

We have shown that the qualitative probability structure is compatible with monotonic belief functions, and a belief structure slightly weaker than that of qualitative belief is compatible with Smets' generalized belief functions. More importantly, the qualitative and quantitative representations of beliefs may lead to the development of a generalized utility theory for decision making with belief functions.

## Acknowledgements

The authors wish to thank P. Smets and Y.T. Hsia for their critical comments and useful suggestions on the axiomatization of belief structures.